\documentclass{article}

\usepackage{arxiv}

\usepackage[utf8]{inputenc} 
\usepackage[T1]{fontenc}    
\usepackage{hyperref}       
\usepackage{url}            
\usepackage{booktabs}       
\usepackage{amsfonts}       
\usepackage{nicefrac}       
\usepackage{microtype}      
\usepackage{lipsum}		
\usepackage{graphicx}
\usepackage{natbib}
\usepackage{doi}
\usepackage{array}
\usepackage{longtable}

\usepackage{tikz}
\usetikzlibrary{shapes,arrows,decorations.pathreplacing}
\usepackage{verbatim}
\usepackage{amsmath}
\newcolumntype{C}[1]{>{\centering\let\newline\\\arraybackslash\hspace{0pt}}m{#1}}

\title{Constructing Multi-label Hierarchical Classification Models for MITRE ATT\&CK Text Tagging}

\author{Andrew Crossman\thanks{Authors listed alphabetically.} \\ JPMorganChase
\And Jonah Dodd \\ JPMorganChase
\And Viralam Ramamurthy Chaithanya Kumar \\ JPMorganChase
\And Riyaz Mohammed \\ JPMorganChase
    \And Andrew R. Plummer \\ JPMorganChase
    \And Chandra Sekharudu  \\ JPMorganChase
    \And Deepak Warrier \\ JPMorganChase
    \And  Mohammad Yekrangian \\ JPMorganChase
}

\date{}

\begin{document}
\maketitle

\begin{abstract}
MITRE ATT\&CK is a cybersecurity knowledge base that organizes threat actor and cyber-attack information into a set of tactics describing the reasons and goals threat actors have for carrying out attacks, with each tactic having a set of techniques that describe the potential methods used in these attacks.  One major application of ATT\&CK is the use of its tactic and technique hierarchy by security specialists as a framework for annotating cyber-threat intelligence reports, vulnerability descriptions, threat scenarios, inter alia, to facilitate downstream analyses.  To date, the tagging process is still largely done manually.  In this technical note, we provide a stratified "task space" characterization of the MITRE ATT\&CK text tagging task for organizing previous efforts toward automation using AIML methods, while also clarifying pathways for constructing new methods.  To illustrate one of the pathways, we use the task space strata to stage-wise construct our own multi-label hierarchical classification models for the text tagging task via experimentation over general cyber-threat intelligence text -- using shareable computational tools and publicly releasing the models to the security community (via https://github.com/jpmorganchase/MITRE\_models).  Our multi-label hierarchical approach yields accuracy scores of roughly $94 \%$ at the tactic level, as well as accuracy scores of roughly $82 \%$ at the technique level (when cast as multiclass transformations).  The models also meet or surpass state-of-the-art performance while relying only on classical machine learning methods -- removing any dependence on LLMs, RAG, agents, or more complex hierarchical approaches (e.g., algorithm adaptation methods, or DAG-based methods).  Moreover, we show that GPT-4o model performance at the tactic level is significantly lower (roughly $60 \%$ accuracy) than our own approach.  We also extend our baseline model to a corpus of threat scenarios for financial applications produced by subject matter experts. 
\end{abstract}


\section{Introduction}

Formal developments in cybersecurity began to take shape throughout the latter half of the 20th century (e.g., packet switching, Diffie-Hellman key exchange, early antivirus software, inter alia) due largely to the dramatic increase in the inter-connectivity of government, industry, and civilian information systems.  The expansion and globalization of the internet and related technologies over the last 30 years necessitated further advancement of security practices (internet security protocols, cloud security protocols, etc.) to protect critical information resources from cyber-attacks and their resulting cost.  Indeed the widening scope of cybersecurity over the last few decades led organizations across government and industry to formulate frameworks that support security specialists in the characterization, detection, mitigation, and prevention of cyber-attacks.    

Lockheed Martin produced a cybersecurity framework focusing on threat-based aspects of risk \citep{hutchins2011intelligence} adapted from the DoD's "kill chain" approach for targeting and engaging an adversary and the DoD's "course of action" model for disruption of adversary activities.  The kill chain approach models the stages of an adversary's progression toward establishing an advanced persistent threat to a system together with potential detection and mitigation capabilities at each stage.  Analysis of multiple kill chains over time yields patterns of behavior, stratified into tactics and techniques, that respectively characterize the "why" and "how" of adversary attacks. 

The MITRE Corporation produces and maintains the ATT\&CK framework \citep{mitre2025} -- a knowledge base constructed from real-world observations of cyber-attacks that includes threat actor groups and their known means of attack. The ATT\&CK framework organizes cyber-attack information hierarchically, with the two main levels being the tactics and techniques that characterize adversarial activities.  In line with the kill chain, the tactics correspond to the "why" of an attack -- indicating the adversary's goals or reasons, while the techniques correspond to "how" the adversaries performed their attacks.  The ATT\&CK framework has been widely adopted and applied within government and industry for threat modeling (see \citealp{11050643} and Section~\ref{sec:mlhm} for specific applications), threat detection and hunting, vulnerability analysis, control validation, as well as broader cybersecurity intelligence analysis.  

Note that the ATT\&CK framework is not tied to a specific model of cyber-attack stages, allowing ATT\&CK to easily function as a cyber-threat intelligence annotation framework.  Indeed, one of the major tasks of security analysts involves reading cyber-threat intelligence reports and tagging the documents and their contents with MITRE ATT\&CK tactics and techniques to facilitate downstream analyses.  To date, the tagging process is still largely carried out by analysts in manual fashion.  Over the last decade a collection of computational methods have taken shape that aim to automate (or semi-automate) the tagging task to reduce analyst toil (see Section~\ref{sec:mitre-tagging} for a review of established and emerging methods).

In this connection, our contributions in this technical note are the following:
\begin{itemize}
    \item A stratified "task space" formulation of the MITRE ATT\&CK text tagging task (Section~\ref{sec:mitre-tagging}) for organizing existing AIML approaches and facilitating further developments;   
    \item A "bottom-up" stage-wise construction of a baseline multi-label hierarchical tagging system for general cyber-intelligence texts following the task space levels  -- the construction process is not beholden to canonical "top-down" AIML modeling approaches or architectures, but rather, organically encapsulates the "Best Practices for MITRE ATT\&CK Mapping" specified in CISA's guide for analysts \citep{cisa2023}, building up from experimentation; 
    \item A comparison of our model performance against GPT-4o during the construction process (Section~\ref{sec:mlhm});
    \item An example of the re-use of the baseline model to bootstrap modeling on new data sets -- using a corpus of threat scenarios for financial applications produced by security specialists within JPMC (Section~\ref{sec:mlhm});
    \item A release of a version of our tagging system that is publicly available for download and use by the security community (via https://github.com/jpmorganchase/MITRE\_models).  
\end{itemize}

\section{MITRE ATT\&CK Text Tagging Task Formulation}  
\label{sec:mitre-tagging}

\begin{table}
\caption{Stratified Task Types for MITRE ATTA\&CK Text Tagging.}
\renewcommand{\arraystretch}{1.9}
\setlength{\tabcolsep}{10pt}
\centering
\begin{tabular}{cC{4cm}cC{4.5cm}c}
    \toprule
    Task ID & Task Type     & \textsc{att\&ck} Mapping Form & Output Details \\ 
    \midrule
    1 & Multiclass Tactic Classification & $D \mapsto T$  &  $T$ is a single tactic selected from a set of tactics   \\
    2 & Multiclass Technique Classification     & $D \mapsto T$ &   $T$ is a single technique selected from a set of techniques \\  
    3 & Multi-label Tactic Classification     &     $D \mapsto \{T_1,\ldots, T_n\}$  &  $T_i$ are tactics  selected from a set of tactics \\  
    4 & Multi-label Technique Classification     &   $D \mapsto \{T_1,\ldots, T_n\}$  &  $T_i$ are techniques  selected from a set of techniques\\  
    5 & Mixed-type Multi-label Classification     &    $D \mapsto \{T_1, \ldots, T_n\}$    &  $T_i$ are tactics and$\slash$or techniques selected from a set of tactics and a set of techniques \\ 
    6 & Multiclass Hierarchical Classification    &    $D \mapsto (T_1, T_2)$    &  $T_1$ is a tactic and $T_2$ is a technique for $T_1$ \\    
    7 & Multi-label Hierarchical Classification    &    $D \mapsto \{T_1,\ldots, T_n\}$    &  $T_i$ is a tuple  $(T^0_i,T^1_i,\ldots, T^k_i)$ where $T^0_i$ is a tactic and each $T^j_i$ is a technique for $T^0_i$ $(j > 0)$ \\
    8 & Text-to-Text Classification     &   $D \mapsto T$    &  $T$ is a text description of tactics or techniques (or both) \\  
    \bottomrule
\end{tabular}
\label{tab:table}
\end{table}

We first provide a general formulation of the MITRE ATT\&CK text tagging task in order to both organize existing work on its (semi-)automation and clarify potential pathways for further development of its AIML-based modeling.  This "task space" formulation, shown in Table~\ref{tab:table},  provides clear strata for such comparisons, while also serving as a road map for "bottom-up" model-building in low-resource-sparse-data settings (as we show in Section~\ref{sec:mlhm}).  Although a comprehensive review of existing work is beyond the scope of this technical note (see \citealp[]{buechel2025sok}, for a recent survey), we follow with a brief AIML-focused review of related works established in the public space that exemplify the different levels of instantiations of the general tagging task (listing references for each type).

The MITRE ATT\&CK text tagging task takes the following general form: 
\begin{center}
$\textsc{att\&ck} : D \mapsto T$
\end{center}
where $D$ is a text document and $T$ is a formal representation of aspects of the ATT\&CK knowledge base (restricted to the Enterprise Matrix v14, herein).  In the simplest form, $D$ is a short document, e.g., a sentence or phrase, and $T$ is a single tactic or technique.  However, both $D$ and $T$ can be made complex.  The input $D$ is extensible to paragraphs, full documents, or even sets of documents, as well as text that varies in topic from general cybersecurity intelligence, to threat scenario descriptions, threat reports, cyber-attack reports, vulnerability descriptions, etc.  (see \citealp{della2025cti} for a review of existing annotated corpora corresponding to the types of $D$, as well as \citealp[]{AlamEtAl2024} for benchmarking).  The formal representations $T$ are extensible from a single tactic or technique to sets of tactics or techniques, sets of tactics together with techniques, textual descriptions of tactics and techniques, hierarchical structures over tactics and techniques, etc.  Task types that capture the nomenclature and descriptions of the common different forms of $T$ are stratified (roughly) in terms of complexity and given IDs in Table~\ref{tab:table}.  We step through descriptions of existing works that exemplify each of these task types below.

Initial efforts to automate the tagging task following the advent of ATT\&CK in 2013 relied primarily on expert-crafted taxonomies and knowledge graphs over ATT\&CK information as a basis for fuzzy string matching algorithms mapping input texts of cyber-threat intel to known graph/taxonomy entries (see MITRE ATT\&CK Extractor, MITRE D3FEND for recent versions, roughly addressing Task IDs 1 and 2 in Table~\ref{tab:table}).  While effective to a degree, the rigidity and inflexibility of these methods (relying on classical NLP-based syntactic and semantic representations of cyber-threat intel entities, relations, and concepts) led to the initiation (and later expansion) of efforts in the emerging machine learning space (e.g., \citealp[]{ayoade2018automated}, \citealp{ampel2021linking}, \citealp{rahman2024alert}).  

The Threat Report ATT\&CK Mapper (TRAM) project began as a cybersecurity community effort to drive machine learning-based progress on the tagging task.  The follow-up TRAM 2.0 (https://github.com/center-for-threat-informed-defense/tram) extended the original project beyond basic machine learning methods via the incorporation of emerging transformer-based text representations known to facilitate tasks akin to ATT\&CK tagging (see also \citealp[]{alves2022leveraging}, \citealp{you2022tim}, \citealp{rani2023ttphunter}, \citealp{rani2024ttpxhunter}), while also adopting a multi-label approach (Task IDs 3 and 4 in Table~\ref{tab:table}, see \citealp[]{mendsaikhan2020automatic}, \citealp{kuppa2021linking}, \citealp{grigorescu2022cve2att} for more on multi-label methods).  Still, the tagging task in TRAM is restricted to just the technique level within the ATT\&CK hierarchy.

In contrast, the Reports Classification by Adversarial Tactics and Techniques (rcATT) system (see \citealp{legoy2020automated} and https://github.com/vlegoy/rcATT) tags cyber-threat intelligence reports with both tactics and techniques (at the document level) using machine learning models trained independently at the tactic and techniques levels of the ATT\&CK framework (falling within Task ID 5 in Table~\ref{tab:table}).  The system is equipped with a UI for recording user feedback to the automated tagging output and using it for updating the tagging models.  The approach, however, does not capture or utilize the known hierarchical relationships between tactics and techniques in the classification process (but rather as a post-classification processing step). 

TTPDrill \citep{10.1145/3134600.3134646} is an ontology-based approach (see  \citealp[]{satvat2021extractor}, \citealp[]{li2022attackg}, and \citealp{alam2023looking} for related approaches) to the tagging task that directly incorporates the hierarchical relationship between tactics and techniques when mapping sentences in cyber-threat intelligence reports  (Task ID 6 in Table~\ref{tab:table}).  Its threat action ontology is manually crafted with fields that hierarchically represent kill chain phases, tactics, and techniques, as well as more specific information on threat action types.  Sentences are mapped to the ontology first through a dependency parser that creates threat action "candidates" from their constituent text, and the candidates are compared against ontology entries via a semantic similarity computation.  The tactic and technique of the best matching ontology entry is assigned to each sentence in a threat report (above a learned threshold).  One drawback is that the mapping produces just one tactic-technique pair for each sentence when multiple such labels may be relevant for cyber-threat analysis.   

A growing number of methods over the last few years attempt to directly address the multi-label nature of the ATT\&CK tagging task on the one hand, in addition to its hierarchical nature on the other (see Task ID 7 in Table~\ref{tab:table}).  These methods incorporate and integrate advances in the general fields of both multi-label classification (e.g., the development of problem transformation versus algorithm adaptation methods, see \citealp{kassim2024multi}) and hierarchical classification (e.g., tree- versus DAG-based methods, see \citealp[]{ramirez2016hierarchical} inter alia) that have taken shape over the last 20 years, cross-cut by concurrent advances in deep learning (see \citealp{LIU2022108826}, \citealp{li2024automated}).  In the next section, we construct our multi-label hierarchical ATT\&CK tagging models, building up along the task space strata in Table~\ref{tab:table}.   

The recent Text-to-Text methods (Task ID 8 in Table~\ref{tab:table}) abstract away from the structure of the multi-label hierarchical characterization of the output of the ATT\&CK tagging task while attempting to preserve and extend the nature of the output.  Specifically, the output of the mapping, $T$, need not be a formal structure, but rather a text that encompasses the information that a multi-label hierarchical structure would contain (e.g., the input document $D$ maps to a set of tactics and/or techniques, while $T$ may also contain additional information).  Many industry and research groups (see \citealp{branescu2024automated}, \citealp[]{fayyazi2024advancing}, \citealp{xu2024intelex}, \citealp{schwartz2025llmcloudhunter}, \citealp{huang2024mitretrieval}, \citealp{nir2025labeling}, \citealp{liu2025cyber}) are moving in the direction of Text-to-Text classification for MITRE ATT\&CK tagging.  While we reserve plans for extension of our multi-label hierarchical approach to Text-to-Text classification, we briefly comment on them in Section~\ref{sec:discussion}.

\begin{figure}
	\centering
	\includegraphics[scale=0.41]{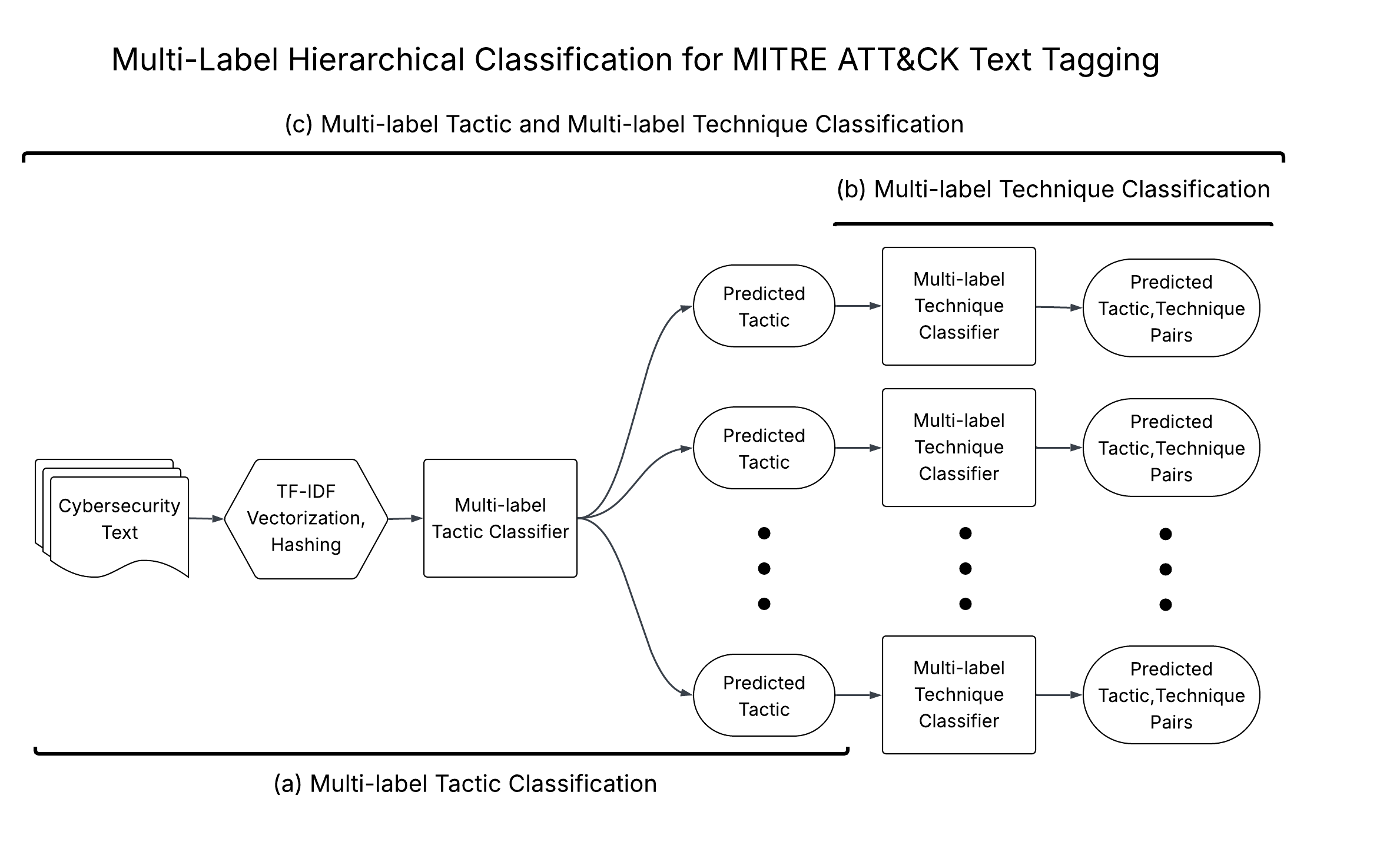}
	\caption{A multi-label hierarchical classification system for the MITRE ATT\&CK text tagging task. Documents are decomposed into sentences that are vectorized using TF-IDF.  The system provides a hashing technique for encrypting the text as a part of the vectorization process.  The first level of hierarchical classification (a) uses a multi-label classification model to predict the top $n$ tactic labels.  The second level (b) uses tactic-specific multi-label classification models, conditioned on the predicted tactics, to provide the top $m$ technique labels for each tactic.  Output for the entire system (c) is a structure of $(n*m)$-many tactic-technique pairs.}
	\label{fig:figml}
\end{figure}

\section{Multi-label Hierarchical ATT\&CK Tagging System Construction and Evaluation}
\label{sec:mlhm}

\begin{figure}
	\centering
	\includegraphics[scale=0.57]{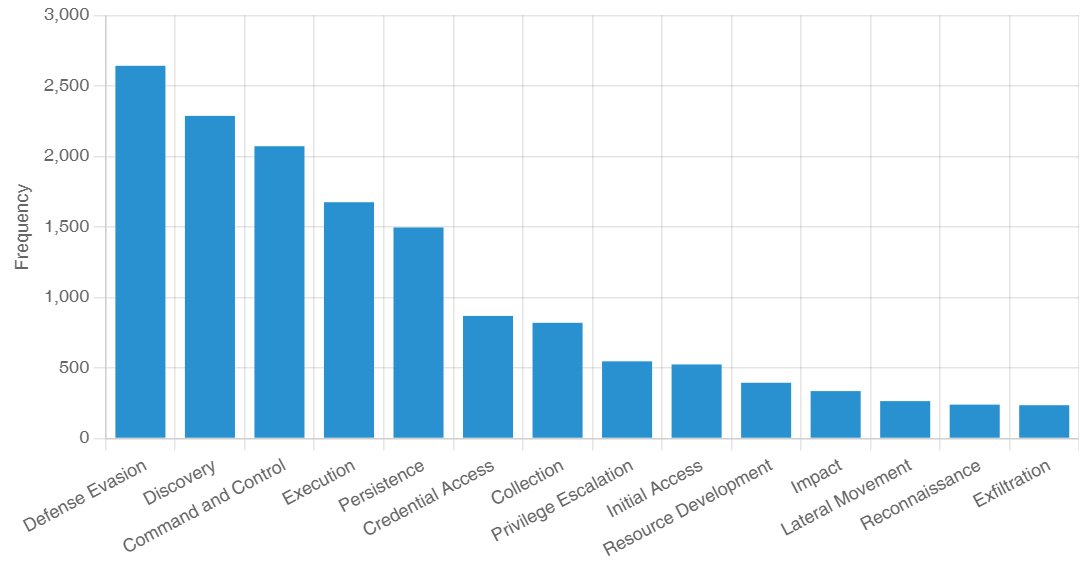}
	\caption{Tactic counts for the baseline cyber-intelligence text data set.  Total (14405), with Defense Evasion (2642), Discovery (2287), Command and Control (2072), Execution (1675), Persistence (1496), Credential Access (869), Collection (820), Privilege Escalation (547), Initial Access (525), Resource Development (395), Impact (336), Lateral Movement (265), Reconnaissance (240), Exfiltration (236).}
	\label{fig:tacticdist}
\end{figure}

Our full multi-label hierarchical classification model architecture for the MITRE ATT\&CK tagging task is shown in Figure~\ref{fig:figml}.  Rather than making a "top-down" architecture selection -- an a priori choice of an AIML modeling architecture -- we took a "bottom-up" sequential approach in building up the architecture, progressing through the task space strata in Table~\ref{tab:table}.  Moreover, we motivated progression through the strata via the results of three experimental stages.  The first two experimental stages rely on a data set of 14405 general cyber-intelligence sentences each of which has a single corresponding gold standard ATT\&CK tactic and technique label.  The data were compiled and curated by cybersecurity specialist within JPMC to ensure data quality.  The distribution of the data set by tactic is shown in Figure~\ref{fig:tacticdist}.  The third experimental stage relies on a second data set of 552 threat scenarios extracted from threat models produced by cybersecurity specialists within JPMC for actual applications used within the bank.  Of these data points, 486 have at least one gold standard tactic label, while 66 have no tactic label (these are omitted from experiments).  Of the 486 data points that have tactic labels, 306 have a single ATT\&CK tactic label\footnote{The tactic distribution is 'Initial Access' (87), 'Impact' (71) 
'Collection' (50), 'Defense Evasion' (29), 'Exfiltration' (29), 'Lateral Movement' (24), 'Privilege Escalation' (20), 'Credential Access' (19), 'Discovery' (15), 'Resource Development' (14), 'Execution' (8), 'Persistence' (7),  'Reconnaissance' (2), 'Command and Control' (1).}.   The remaining 180 data points are multi-labeled.  We note the class imbalances and sparsity in the data sets and leave them as is to better replicate real-world data conditions. Moreover, supporting experiments that accounted for the class imbalances showed similar results to our main experiments.  

\textbf{Experimental Stage 1} -- We conducted a pilot study comparing the performance of a stochastic gradient descent support vector machine (from scikit-learn, referred to as our baseline multiclass SGD model in the remainder of this paper) against GPT-4o in multiclass tactic classification (Task ID 1 in Table~\ref{tab:table}).  Specifically, we start with 
\begin{center}
$\textsc{att\&ck} : D \mapsto T$
\end{center}
limiting $T$ to a single ATT\&CK tactic and restricting $D$ to sentences.  The multiclass SGD model was selected based on earlier experimental results showing that the model type outperformed other standard machine learning models on this same multiclass task.  Full parameter details of the model will accompany our public release.  The selection of GPT-4o for comparison was due to its being the latest release available to us when initiating our experiments.  The temperature is left at the default setting of 1.

\begin{table}
\caption{Classification evaluation results for ATT\&CK tactic tagging pilot study.  Results show that a multiclass SGD model generally outperforms GPT-4o in multiclass prediction for a given input cyber-intelligence sentence.}
\setlength{\tabcolsep}{10pt}
\centering
\begin{tabular}{lC{4cm}C{5cm}}
\toprule
Evaluation Attribute & Multiclass SGD Model & GPT-4o \\
\midrule
Accuracy & \textbf{0.8195} & 0.59  \\
F1       & \textbf{0.7795} & 0.60 \\
\midrule
Accuracy parsed by Tactic & & \\
\midrule
Defense evasion & \textbf{0.8272} & 0.6345  \\
Discovery & \textbf{0.8969} & 0.6433  \\
Persistence & \textbf{0.7903} & 0.5017  \\
Initial access & \textbf{0.7307} & 0.6286  \\
Collection & \textbf{0.8424} & 0.6402  \\
Execution & \textbf{0.8055} & 0.5194  \\
Lateral movement & 0.5957 & \textbf{0.6226}  \\
Impact & \textbf{0.7142} & 0.6716  \\
Command and control & \textbf{0.8561} & 0.5783  \\
Credential access & 0.7621 & \textbf{0.8046}  \\
Privilege escalation & \textbf{0.7130} & 0.2091  \\
Reconnaissance & 0.6388 & \textbf{0.6875} \\
Resource development & \textbf{0.8117} & 0.5190  \\
Exfiltration & 0.5813 & \textbf{0.7234}  \\
\bottomrule
\end{tabular}
\label{tab:tabllm}
\end{table}

The cyber-intelligence data was randomly split into a training set (80\%) and a test set (20\%) ensuring faithful representation of the tactic distribution within each.  The multiclass SGD model was trained on the former set and evaluated on the later.  All of the textual input to the multiclass SGD model was first transformed into vector representations, in this case, TF-IDF for simplicity.  The GPT-4o model was evaluated on the test set by saturating the prompt below with the test sentences, one at a time, in their textual rather than vectorized forms. 
\begin{quote}
\begin{verbatim}
Look at this cyber-intelligence text and label it with a mitre tag 
from the selection provided to you in this message.

RETURN YOUR RESPONSE IN THE FOLLOWING JSON FORMAT WITHOUT MARKDOWN:
{{
    "Tag": "YOUR MITRE TAG"
}}

IT IS EXTREMELY IMPORTANT THAT YOU RETURN THE EXACT "NAME" VALUE 
FOR A MAXIMUM REWARD.

MITRE_TAGS:
    * TA0006 - Credential Access * TA0002 - Execution  * TA0003 - Persistence 
    * TA0001 - Initial Access * TA0005 - Defense Evasion * TA0007 - Discovery
    * TA0008 - Lateral Movement * TA0009 - Collection * TA0010 - Exfiltration
    * TA0043 - Reconnaissance * TA0040 - Impact * TA0042 - Resource Development
    * TA0011 - Command and Control * TA0004 - Privilege Escalation

cyber-intelligence text: 
{input-sentence} 
\end{verbatim}
\end{quote}
Note that this GPT-4o tagging approach is technically Text-to-Text classification (Task ID 8 in Table~\ref{tab:table}), necessitating that its output be normalized to ensure that generated tactic labels were comparable to the ground truth tactic labels.  Results in Table~\ref{tab:tabllm} show that our multiclass SGD model significantly outperformed GPT-4o over cyber-intelligence data at the tactic level.  Given the results, and the overall light footprint, share-ability, and extensibility of the multiclass SGD model, we took it as the point of departure for our multi-label hierarchical classification system.

\textbf{Experimental Stage 2} -- We next conducted a set of experiments to address three goals.  The first goal concerned the "problem transformation" for the multiclass SGD model, that is, making it behave more like a multi-label classification model (moving up to Task ID 3 in Table~\ref{tab:table}).  The second goal involved extending the transformed multi-label SGD model to classification at the technique level, i.e., making it properly multi-label and hierarchical (moving up to Task ID 7 in Table~\ref{tab:table}).  The third goal concerned ensuring the safety of data used to train the models in service of public release, while not impeding model performance.  

\begin{table}
\caption{Multi-label classification evaluation results on the cyber-intelligence baseline data set.  Results show that when adopting a top-$n$ labeling method performance increases substantially at the tactic level.  Moreover, the data hashing for security does not impact model performance.}
\setlength{\tabcolsep}{10pt}
\centering
\begin{tabular}{lC{4cm}C{5cm}}
\toprule
Evaluation Attribute & Multi-label Classifier using Multiclass SGD Models & Multi-label Classifier using Multiclass SGD Models (Hashing) \\
\midrule
Top $n=3$ Accuracy & 0.8264 & 0.8105 \\
Tactic accuracy & 0.9455 & 0.9427  \\
Technique accuracy & 0.8264 & 0.8105  \\
Tactics correct & 2724 & 2716 \\
Techniques correct & 2381 & 2335  \\
Both correct predictions & 2381 & 2335  \\
Total predictions & 2881 & 2881 \\
\midrule
Top $n=3$ Accuracy parsed by Tactic & & \\
\midrule
Defense evasion & 0.9424 & 0.9597  \\
Discovery & 0.9635 & 0.9700  \\
Persistence & 0.9537 & 0.9466  \\
Initial access & 0.8942 & 0.8654  \\
Collection & 0.9576 & 0.9455  \\
Execution & 0.9444 & 0.9306  \\
Lateral movement & 0.8723 & 0.8511  \\
Impact & 0.9107 & 0.8750  \\
Command and control & 0.9688 & 0.9736  \\
Credential access & 0.9351 & 0.9297  \\
Privilege escalation & 0.9130 & 0.9043  \\
Reconnaissance & 0.8333 & 0.8056 \\
Resource development & 0.9647 & 0.9412  \\
Exfiltration & 0.9070 & 0.8605  \\
\bottomrule
\end{tabular}
\label{tab:tab2}
\end{table}

For the first goal, we simply modify the output of the multiclass SGD model to be the top $n$ predicted tactics (choosing $n=3$), rather than the top 1 tactic (corresponding to (a) in Figure~\ref{fig:figml} with $n=3$).  The performance of this multi-label ATT\&CK tagging system is measured in terms of a standard subset operation.  That is, suppose we are given an input sentence $S$ with a ground truth tactic $T$, and let $\{T_1,T_2,T_3\}$ be the top 3 multi-label SGD tactic prediction.  The prediction for $S$ is considered correct if and only if $\{T\} \subseteq \{T_1,T_2,T_3\}$ (known formally as top-$n$ accuracy, where $n=3$).  The multiclass SGD model was trained from scratch as a multiclass model over the cyber-intelligence data set, but then evaluated using the-multi-label accuracy method.  System performance for the multi-label accuracy evaluation at the tactic level is shown in the top partition of Table~\ref{tab:tab2}, with accuracy reaching 94\%, parsed out by tactic in the lower partition.  While the boost in performance for the SGD model is expected with the more general formulation of accuracy relative to Experimental Stage 1,  the  improvement in results parsed out by tactic support the treatment of the tagging task as multi-label rather than multiclass. 

For the second goal, we extend our tactic-level SGD model to the technique level by simply training multiclass SGD classifiers for the techniques associated with each tactic.  That is, we first parse the cyber-intelligence data into tactic-specific data sets, and then we train tactic-specific multiclass SGD models to make multiclass predictions over the techniques for that tactic using a randomized 80-20 training-test split of the corresponding tactic-specific data sets.  The multi-label mapping at the technique level is again based on top 3 multiclass SGD classifier output (corresponding to (b) in Figure~\ref{fig:figml} with $m=3$).  The final system prediction for a given input sentence is three tactics, each of which is paired with three techniques (corresponding to (c) in Figure~\ref{fig:figml} with $n=m=3$).  Assuming the restriction of top $n=3$ predictions at both levels, multi-label hierarchical system accuracy is defined as follows.  Let $S$ be an input sentence with ground truth tactic and technique labels $(Ta_S, Te_S)$.  Let $\{T_1,T_2,T_3\}$ be the top 3 multi-label SGD tactic prediction and for each $T_i$ let $\{T^1_i, T^2_i, T^3_i \}$ be the technique predictions that follow.  The tactic-technique predictions are arranged into a set of nine pairs $\{(T_i, T^j_i) \mid\text{ for } 1\le i, j \le 3 \}$.   The prediction for $S$ is considered correct if and only if $ \{(Ta_S, Te_S)\} \subseteq \{(T_i, T^j_i) \mid\text{ for } 1\le i, j \le 3 \}$.  Overall accuracy of the multi-label hierarchical system is shown in the top partition of Table~\ref{tab:tab2}, reaching 82\%.  Moreover, the table shows that techniques are never predicted correctly together with an incorrect tactic prediction (that is, "Techniques correct" is the same as "Both correct"), showing the merit of the hierarchical approach.  Specifically, the proper DAG structure of the MITRE ATT\&CK hierarchy (a technique can have multiple tactic parents) can be dealt with using multi-label hierarchical modeling. 

For the third goal, the system includes a hashing option that encrypts the data used in SGD model training as a part of the vectorization process.  We tested our hashing option using MurmurHash3, though others are available through scikit-learn. The hashed representations were also run through TfidfTransformer to ensure IDF weighting.  We trained two multiclass SGD models from scratch at the tactic level (corresponding to (a) in Figure~\ref{fig:figml} with $n=3$) using the cyber-intelligence data, one exposed to the standard TF-IDF vectors and the other exposed to the hashing-based vectors.  Both models were evaluated using the multi-label accuracy method for the tactic level described above.  Results of the overall comparison are shown in the top partition of Table~\ref{tab:tab2},  parsed out at the tactic level in the lower partition of Table~\ref{tab:tab2}.  Note that the encryption method does not significantly impact system performance.   This allows us to share the models out to the community with a high degree of security on the sensitive data used to train the models.

\textbf{Experimental Stage 3} -- Our final set of experiments is two-fold, investigating how well the multi-label SGD models worked on new data sets that differ in content from the general cyber-intelligence data on the one hand, and how well the "problem transformed" multi-label SGD model type would perform on data points with actual gold standard multi-labels.  We note that both experiments were carried out on the threat scenario data set, which contains only 486 data points, sparse category counts, and labels only at the tactic level -- limiting the interpretation of the results.  Moreover, since the data set contains true multi-label data points, we extend the multi-label accuracy definition for tactics from Experimental Stage 2 as follows.  Suppose we are given an input sentence $S$ with ground truth tactics $\{T^S_1, T^S_2,\ldots,T^S_n \}$, and let $\{T_1,T_2,T_3\}$ be the top 3 multi-label SGD model tactic prediction.  The number of correct predictions for $S$ is the cardinality of the set intersection $\{T^S_1, T^S_2,\ldots,T^S_n \} \cap \{T_1,T_2,T_3\}$.  This formulation limits the number of correct multi-label predictions to three for each sentence $S$, however there are only seven data points with four or more multi-labels, so impact on performance is minimal.  

\begin{table}
\caption{Multi-label Tactic-Level Classification Evaluation Results for Threat Scenarios.  Results show that the baseline Multi-label SGD trained on cyber-intelligence data does not immediately generalize to threat scenario tagging, However, the underlying model architecture is adaptable showing improvement with only a small amount of training data.}
\setlength{\tabcolsep}{10pt}
\centering
\begin{tabular}{lC{4cm}C{5cm}}
\toprule
Evaluation Attribute & Multi-label Classifier using Multiclass SGD Models Trained on Cyber-intel Data & Multi-label Classifier using Multiclass SGD Models trained on Threat Scenario Data \\
\midrule
Top $n=3$ Accuracy & 0.41 & \textbf{0.66} \\
Tactics correct & 54 &  \textbf{88} \\
Total predictions & 132 & 132 \\
\midrule
Top $n=3$ Accuracy parsed by Tactic & & \\
\midrule
Defense evasion & 0.5 & \textbf{0.62}  \\
Discovery & 0.42 & \textbf{0.57}  \\
Persistence & \textbf{0.75} & 0.25  \\
Initial access & 0.19 & \textbf{0.74}  \\
Collection & 0.60 & \textbf{0.75}  \\
Execution & 0.00 & 0.00  \\
Lateral movement & 0.16 & 0.16  \\
Impact & 0.54 & \textbf{0.87}  \\
Command and control & 0.00 & \textbf{1.00} \\
Credential access & 0.70 & 0.70  \\
Privilege escalation & 0.00 & \textbf{0.40}  \\
Reconnaissance & 0.00 & 0.00 \\
Resource development & 0.00 & \textbf{0.50}  \\
Exfiltration & 0.62 & 0.62  \\
\bottomrule
\end{tabular}
\label{tab:tab4}
\end{table}

The threat scenario data was randomly split into a training set ($\sim$80\%) and a test set ($\sim$20\%) ensuring faithful representation of the tactic distribution within each as best as possible given the data sparsity.  The split yielded a test set consisting of 111 threat scenario sentences with a grand total of 132 tactic ground truth labels (due to the test set containing multi-labeled threat scenarios).  All of the textual data was again transformed into TF-IDF vector representations.  In the first experiment, the threat scenario test set was simply run through the multiclass SGD model with tactic prediction accuracy computed using the defined set intersection cardinality.  For the second experiment, a multiclass SGD model was trained from scratch as a multiclass classifier on the training set and then evaluated as a multi-label model on the test set using the defined set intersection accuracy.  Results in Table~\ref{tab:tab4} show that the baseline Multi-label SGD trained on cyber-intelligence data does not immediately generalize to threat scenario tagging, However, the underlying model architecture is adaptable, showing improvement with only a small amount of training data.  While more exploration is needed, the approach is in line with low-resource-sparse-data model building.

\section{Review}
\label{sec:discussion}

In this technical note, we began by providing a general "task space" formulation of the MITRE ATT\&CK text tagging task for organizing existing AIML-related work and facilitating further developments.  The formulation gave structure to our "bottom-up" stage-wise construction of a baseline multi-label hierarchical tagging system for general cyber-intelligence texts, as we leveled up through the task space strata based on experimental results.  Our system construction process eschewed the canonical "top-down" AIML modeling predispositions in favor of incorporating the "Best Practices for MITRE ATT\&CK Mapping" specified in CISA's guide for analysts \citep{cisa2023}.  During our system build-up we showed that our baseline models outperformed GPT-4o on multiclass tactic prediction.  We also showed how to re-use the baseline models to bootstrap modeling on new data sets -- exemplifying this re-use on a set of threat scenarios for financial applications produced by security specialists within JPMC.  We also implemented a model-performance-preserving hashing method in supporting our public release of a tagging system for download and use by the security community.  

We close with two main observations that came to light in producing this technical note.  The first is, there are a great many approaches to the MITRE ATT\&CK text tagging task.  Yet wide-spread adoption of any of these approaches by security specialists seems to be rare, if not nonexistent.  Usability of the models and systems in low-resource-sparse-data settings (especially, if customizable) may likely be a prerequisite to system adoption, even when the more advanced approaches have higher performance scores.  Hence the reason for the public release of our system, which is straightforward to set up and use.  The second observation is, the cybersecurity community is deeply interested in technological advancements, as they both impact and facilitate cybersecurity activities.  Yet, to a large degree there is a gap between the highly specialized activities of this community and a rich mathematical/technical literature, in AIML and beyond, than can benefit community efforts.  Hence our rigorous formulation of the ATT\&CK tagging task and mapping to existing cybersecurity works.  We aim to further bridge this gap in later publications.

\section{Acknowledgments}

We thank the JPMorganChase Cybersecurity Community and appreciate your contributions and feedback.  

This paper was prepared for informational purposes with contributions from the Cybersecurity and Technology Controls organization of JPMorgan Chase \& Co. This paper is not a product of the Research Department of JPMorgan Chase \& Co. or its affiliates. Neither JPMorgan Chase \& Co. nor any of its affiliates makes any explicit or implied representation or warranty and none of them accept any liability in connection with this paper, including, without limitation, with respect to the completeness, accuracy, or reliability of the information contained herein and the potential legal, compliance, tax, or accounting effects thereof. This document is not intended as investment research or investment advice, or as a recommendation, offer, or solicitation for the purchase or sale of any security, financial instrument, financial product or service, or to be used in any way for evaluating the merits of participating in any transaction.

\bibliographystyle{unsrtnat}
\bibliography{references}

\begin{thebibliography}{33}
\providecommand{\natexlab}[1]{#1}
\providecommand{\url}[1]{\texttt{#1}}
\expandafter\ifx\csname urlstyle\endcsname\relax
  \providecommand{\doi}[1]{doi: #1}\else
  \providecommand{\doi}{doi: \begingroup \urlstyle{rm}\Url}\fi

\bibitem[Hutchins et~al.(2011)Hutchins, Cloppert, Amin, et~al.]{hutchins2011intelligence}
Eric~M Hutchins, Michael~J Cloppert, Rohan~M Amin, et~al.
\newblock Intelligence-driven computer network defense informed by analysis of adversary campaigns and intrusion kill chains.
\newblock \emph{Leading Issues in Information Warfare \& Security Research}, 1\penalty0 (1):\penalty0 80, 2011.

\bibitem[{The MITRE Corporation}(2025)]{mitre2025}
{The MITRE Corporation}.
\newblock {MITRE ATT\&CK}, 2025.
\newblock URL \url{https://attack.mitre.org/}.

\bibitem[Crossman et~al.(2025)Crossman, Plummer, Sekharudu, Warrier, and Yekrangian]{11050643}
Andrew Crossman, Andrew~R. Plummer, Chandra Sekharudu, Deepak Warrier, and Mohammad Yekrangian.
\newblock Auspex: Building threat modeling tradecraft into an artificial intelligence-based copilot.
\newblock In \emph{2025 IEEE Conference on Artificial Intelligence (CAI)}, pages 1160--1167, 2025.
\newblock \doi{10.1109/CAI64502.2025.00201}.

\bibitem[{Cybersecurity and Infrastructure Security Agency}(2023)]{cisa2023}
{Cybersecurity and Infrastructure Security Agency}.
\newblock {Best Practices for MITRE ATT\&CK Mapping}, 2023.
\newblock URL \url{https://www.cisa.gov/sites/default/files/2023-01/Best%20Practices%20for%20MITRE%20ATTCK%20Mapping.pdf}.

\bibitem[Büchel et~al.(2025)Büchel, Paladini, Longari, Carminati, Zanero, Binyamini, Engelberg, Klein, Guizzardi, Caselli, Continella, van Steen, Peter, and van Ede]{buechel2025sok}
Marvin Büchel, Tommaso Paladini, Stefano Longari, Michele Carminati, Stefano Zanero, Hodaya Binyamini, Gal Engelberg, Dan Klein, Giancarlo Guizzardi, Marco Caselli, Andrea Continella, Maarten van Steen, Andreas Peter, and Thijs van Ede.
\newblock Sok: Automated ttp extraction from cti reports – are we there yet?
\newblock In \emph{34th USENIX Security Symposium}, Seattle, WA, USA, 2025.
\newblock URL \url{https://www.usenix.org/conference/usenixsecurity25/presentation/buechel}.

\bibitem[Della~Penna et~al.(2025)Della~Penna, Natella, Orbinato, Parracino, and Pianese]{della2025cti}
Sofia Della~Penna, Roberto Natella, Vittorio Orbinato, Lorenzo Parracino, and Luciano Pianese.
\newblock Cti-hal: A human-annotated dataset for cyber threat intelligence analysis.
\newblock \emph{arXiv preprint arXiv:2504.05866}, 2025.

\bibitem[Alam et~al.(2024)Alam, Bhusal, Nguyen, and Rastogi]{AlamEtAl2024}
Md~Tanvirul Alam, Dipkamal Bhusal, Le~Nguyen, and Nidhi Rastogi.
\newblock Ctibench: A benchmark for evaluating llms in cyber threat intelligence.
\newblock In \emph{Advances in Neural Information Processing Systems 37}. NeurIPS, 2024.

\bibitem[Ayoade et~al.(2018)Ayoade, Chandra, Khan, Hamlen, and Thuraisingham]{ayoade2018automated}
Gbadebo Ayoade, Swarup Chandra, Latifur Khan, Kevin Hamlen, and Bhavani Thuraisingham.
\newblock Automated threat report classification over multi-source data.
\newblock In \emph{2018 IEEE 4th International Conference on Collaboration and Internet Computing (CIC)}, pages 236--245. IEEE, 2018.

\bibitem[Ampel et~al.(2021)Ampel, Samtani, Ullman, and Chen]{ampel2021linking}
Benjamin Ampel, Sagar Samtani, Steven Ullman, and Hsinchun Chen.
\newblock Linking common vulnerabilities and exposures to the mitre att\&ck framework: A self-distillation approach.
\newblock \emph{arXiv preprint arXiv:2108.01696}, 2021.

\bibitem[Rahman et~al.(2024)Rahman, Halim, Singhal, and Khan]{rahman2024alert}
Fariha~Ishrat Rahman, Sadaf~Md Halim, Anoop Singhal, and Latifur Khan.
\newblock Alert: A framework for efficient extraction of attack techniques from cyber threat intelligence reports using active learning.
\newblock In \emph{IFIP Annual Conference on Data and Applications Security and Privacy}, pages 203--220. Springer, 2024.

\bibitem[Alves et~al.(2022)Alves, Geraldo~Filho, and Gon{\c{c}}alves]{alves2022leveraging}
Paulo~MMR Alves, PR~Geraldo~Filho, and Vin{\'\i}cius~P Gon{\c{c}}alves.
\newblock Leveraging bert's power to classify ttp from unstructured text.
\newblock In \emph{2022 Workshop on Communication Networks and Power Systems (WCNPS)}, pages 1--7. IEEE, 2022.

\bibitem[You et~al.(2022)You, Jiang, Jiang, Yang, Liu, Feng, Wang, and Li]{you2022tim}
Yizhe You, Jun Jiang, Zhengwei Jiang, Peian Yang, Baoxu Liu, Huamin Feng, Xuren Wang, and Ning Li.
\newblock Tim: threat context-enhanced ttp intelligence mining on unstructured threat data.
\newblock \emph{Cybersecurity}, 5\penalty0 (1):\penalty0 3, 2022.

\bibitem[Rani et~al.(2023)Rani, Saha, Maurya, and Shukla]{rani2023ttphunter}
Nanda Rani, Bikash Saha, Vikas Maurya, and Sandeep~Kumar Shukla.
\newblock Ttphunter: Automated extraction of actionable intelligence as ttps from narrative threat reports.
\newblock In \emph{Proceedings of the 2023 australasian computer science week}, pages 126--134, 2023.

\bibitem[Rani et~al.(2024)Rani, Saha, Maurya, and Shukla]{rani2024ttpxhunter}
Nanda Rani, Bikash Saha, Vikas Maurya, and Sandeep~Kumar Shukla.
\newblock Ttpxhunter: Actionable threat intelligence extraction as ttps from finished cyber threat reports.
\newblock \emph{Digital Threats: Research and Practice}, 5\penalty0 (4):\penalty0 1--19, 2024.

\bibitem[Mendsaikhan et~al.(2020)Mendsaikhan, Hasegawa, Yamaguchi, and Shimada]{mendsaikhan2020automatic}
Otgonpurev Mendsaikhan, Hirokazu Hasegawa, Yukiko Yamaguchi, and Hajime Shimada.
\newblock Automatic mapping of vulnerability information to adversary techniques.
\newblock In \emph{The Fourteenth International Conference on Emerging Security Information, Systems and Technologies SECUREWARE2020}, 2020.

\bibitem[Kuppa et~al.(2021)Kuppa, Aouad, and Le-Khac]{kuppa2021linking}
Aditya Kuppa, Lamine Aouad, and Nhien-An Le-Khac.
\newblock Linking cve’s to mitre att\&ck techniques.
\newblock In \emph{Proceedings of the 16th International Conference on Availability, Reliability and Security}, pages 1--12, 2021.

\bibitem[Grigorescu et~al.(2022)Grigorescu, Nica, Dascalu, and Rughinis]{grigorescu2022cve2att}
Octavian Grigorescu, Andreea Nica, Mihai Dascalu, and Razvan Rughinis.
\newblock Cve2att\&ck: Bert-based mapping of cves to mitre att\&ck techniques.
\newblock \emph{Algorithms}, 15\penalty0 (9):\penalty0 314, 2022.

\bibitem[Legoy et~al.(2020)Legoy, Caselli, Seifert, and Peter]{legoy2020automated}
Valentine Legoy, Marco Caselli, Christin Seifert, and Andreas Peter.
\newblock Automated retrieval of att\&ck tactics and techniques for cyber threat reports.
\newblock \emph{arXiv preprint arXiv:2004.14322}, 2020.

\bibitem[Husari et~al.(2017)Husari, Al-Shaer, Ahmed, Chu, and Niu]{10.1145/3134600.3134646}
Ghaith Husari, Ehab Al-Shaer, Mohiuddin Ahmed, Bill Chu, and Xi~Niu.
\newblock Ttpdrill: Automatic and accurate extraction of threat actions from unstructured text of cti sources.
\newblock In \emph{Proceedings of the 33rd Annual Computer Security Applications Conference}, ACSAC '17, page 103–115, New York, NY, USA, 2017. Association for Computing Machinery.
\newblock ISBN 9781450353458.
\newblock \doi{10.1145/3134600.3134646}.
\newblock URL \url{https://doi.org/10.1145/3134600.3134646}.

\bibitem[Satvat et~al.(2021)Satvat, Gjomemo, and Venkatakrishnan]{satvat2021extractor}
Kiavash Satvat, Rigel Gjomemo, and VN~Venkatakrishnan.
\newblock Extractor: Extracting attack behavior from threat reports.
\newblock In \emph{2021 IEEE European Symposium on Security and Privacy (EuroS\&P)}, pages 598--615. IEEE, 2021.

\bibitem[Li et~al.(2022)Li, Zeng, Chen, and Liang]{li2022attackg}
Zhenyuan Li, Jun Zeng, Yan Chen, and Zhenkai Liang.
\newblock Attackg: Constructing technique knowledge graph from cyber threat intelligence reports.
\newblock In \emph{European Symposium on Research in Computer Security}, pages 589--609. Springer, 2022.

\bibitem[Alam et~al.(2023)Alam, Bhusal, Park, and Rastogi]{alam2023looking}
Md~Tanvirul Alam, Dipkamal Bhusal, Youngja Park, and Nidhi Rastogi.
\newblock Looking beyond iocs: Automatically extracting attack patterns from external cti.
\newblock In \emph{Proceedings of the 26th international symposium on research in attacks, intrusions and defenses}, pages 92--108, 2023.

\bibitem[Kassim et~al.(2024)Kassim, Viktor, and Michalowski]{kassim2024multi}
Mohammed~Awal Kassim, Herna Viktor, and Wojtek Michalowski.
\newblock Multi-label lifelong machine learning: A scoping review of algorithms, techniques, and applications.
\newblock \emph{IEEE Access}, 12:\penalty0 74539--74557, 2024.

\bibitem[Ram{\'\i}rez-Corona et~al.(2016)Ram{\'\i}rez-Corona, Sucar, and Morales]{ramirez2016hierarchical}
Mallinali Ram{\'\i}rez-Corona, L~Enrique Sucar, and Eduardo~F Morales.
\newblock Hierarchical multilabel classification based on path evaluation.
\newblock \emph{International Journal of Approximate Reasoning}, 68:\penalty0 179--193, 2016.

\bibitem[Liu et~al.(2022)Liu, Wang, and Chen]{LIU2022108826}
Chenjing Liu, Junfeng Wang, and Xiangru Chen.
\newblock Threat intelligence att\&ck extraction based on the attention transformer hierarchical recurrent neural network.
\newblock \emph{Applied Soft Computing}, 122:\penalty0 108826, 2022.
\newblock ISSN 1568-4946.
\newblock \doi{https://doi.org/10.1016/j.asoc.2022.108826}.
\newblock URL \url{https://www.sciencedirect.com/science/article/pii/S1568494622002289}.

\bibitem[Li et~al.(2024)Li, Huang, and Chen]{li2024automated}
Lingzi Li, Cheng Huang, and Junren Chen.
\newblock Automated discovery and mapping att\&ck tactics and techniques for unstructured cyber threat intelligence.
\newblock \emph{Computers \& Security}, 140:\penalty0 103815, 2024.

\bibitem[Branescu et~al.(2024)Branescu, Grigorescu, and Dascalu]{branescu2024automated}
Ioana Branescu, Octavian Grigorescu, and Mihai Dascalu.
\newblock Automated mapping of common vulnerabilities and exposures to mitre att\&ck tactics.
\newblock \emph{Information}, 15\penalty0 (4):\penalty0 214, 2024.

\bibitem[Fayyazi et~al.(2024)Fayyazi, Taghdimi, and Yang]{fayyazi2024advancing}
Reza Fayyazi, Rozhina Taghdimi, and Shanchieh~Jay Yang.
\newblock Advancing ttp analysis: Harnessing the power of large language models with retrieval augmented generation.
\newblock In \emph{2024 Annual Computer Security Applications Conference Workshops (ACSAC Workshops)}, pages 255--261. IEEE, 2024.

\bibitem[Xu et~al.(2024)Xu, Wang, Liu, Lin, Liu, Lim, and Dong]{xu2024intelex}
Ming Xu, Hongtai Wang, Jiahao Liu, Yun Lin, Chenyang Xu~Yingshi Liu, Hoon~Wei Lim, and Jin~Song Dong.
\newblock Intelex: A llm-driven attack-level threat intelligence extraction framework.
\newblock \emph{arXiv preprint arXiv:2412.10872}, 2024.

\bibitem[Schwartz et~al.(2025)Schwartz, Ben-Shimol, Mimran, Elovici, and Shabtai]{schwartz2025llmcloudhunter}
Yuval Schwartz, Lavi Ben-Shimol, Dudu Mimran, Yuval Elovici, and Asaf Shabtai.
\newblock Llmcloudhunter: Harnessing llms for automated extraction of detection rules from cloud-based cti.
\newblock In \emph{Proceedings of the ACM on Web Conference 2025}, pages 1922--1941, 2025.

\bibitem[Huang et~al.(2024)Huang, Vaitheeshwari, Chen, Lin, Hwang, Lin, Lai, Wu, Chen, Liao, et~al.]{huang2024mitretrieval}
Yi-Ting Huang, R~Vaitheeshwari, Meng-Chang Chen, Ying-Dar Lin, Ren-Hung Hwang, Po-Ching Lin, Yuan-Cheng Lai, Eric Hsiao-Kuang Wu, Chung-Hsuan Chen, Zi-Jie Liao, et~al.
\newblock Mitretrieval: Retrieving mitre techniques from unstructured threat reports by fusion of deep learning and ontology.
\newblock \emph{IEEE Transactions on Network and Service Management}, 21\penalty0 (4):\penalty0 4871--4887, 2024.

\bibitem[Nir et~al.(2025)Nir, Kaiser, Giladi, Sharabi, Moyal, Shpolyansky, Murillo, Elyashar, and Puzis]{nir2025labeling}
Daniel Nir, Florian~Klaus Kaiser, Shay Giladi, Sapir Sharabi, Raz Moyal, Shalev Shpolyansky, Andres Murillo, Aviad Elyashar, and Rami Puzis.
\newblock Labeling network intrusion detection system (nids) rules with mitre att\&ck techniques: Machine learning vs. large language models.
\newblock \emph{Big Data and Cognitive Computing}, 9\penalty0 (2):\penalty0 23, 2025.

\bibitem[Liu et~al.(2025)Liu, Liang, Yan, Jang, Mao, Ye, Jia, and Xi]{liu2025cyber}
Xiaoqun Liu, Jiacheng Liang, Qiben Yan, Jiyong Jang, Sicheng Mao, Muchao Ye, Jinyuan Jia, and Zhaohan Xi.
\newblock {CyLens: Towards Reinventing Cyber Threat Intelligence in the Paradigm of Agentic Large Language Models}.
\newblock \emph{arXiv preprint arXiv:2502.20791}, 2025.

\end{thebibliography}

\end{document}